\documentclass[sigconf,nonacm]{acmart}

\usepackage{multirow}
\usepackage{graphicx}
\usepackage{caption}
\usepackage{subcaption}

\AtBeginDocument{%
  \providecommand\BibTeX{{%
    \normalfont B\kern-0.5em{\scshape i\kern-0.25em b}\kern-0.8em\TeX}}}

\setcopyright{acmcopyright}
\copyrightyear{2023}
\acmYear{2023}
\acmDOI{XXXXXXX.XXXXXXX}

\acmConference[JCDL '23]{ACM/IEEE-CS Joint Conference on Digital Libraries}{June 26--30, 2023}{New Mexico, USA}
\acmBooktitle{JCDL '23: ACM/IEEE-CS Joint Conference on Digital Libraries, June 26--30, 2023, New Mexico, USA} 
\acmPrice{15.00}
\acmISBN{978-1-4503-XXXX-X/18/06}

\begin{document}

\title{\textsc{CitePrompt}: Using Prompts to Identify Citation Intent in Scientific Papers}

\author{Avishek Lahiri}
\affiliation{%
  \institution{Indian Association for the \\Cultivation of Science}
  \streetaddress{Jadavpur}
  \city{Kolkata}
  \state{West Bengal}
  \country{India}
  \postcode{700032}
}
\email{avisheklahiri2014@gmail.com}

\author{Debarshi Kumar Sanyal}
\affiliation{%
  \institution{Indian Association for the \\Cultivation of Science}
  \streetaddress{Jadavpur}
  \city{Kolkata}
  \state{West Bengal}
  \country{India}
  \postcode{700032}
}
\email{debarshi.sanyal@iacs.res.in}

\author{Imon Mukherjee}
\affiliation{%
  \institution{Indian Institute of \\Information Technology}
  \streetaddress{Kalyani}
  \city{Nadia}
  \state{West Bengal}
  \country{India}}
\email{imon@iiitkalyani.ac.in}

\renewcommand{\shortauthors}{Lahiri, Sanyal, and Mukherjee}

\begin{abstract}
  Citations in scientific papers not only help us trace the intellectual lineage but also are a useful indicator of the scientific significance of the work. Citation intents prove beneficial as they specify the role of the citation in a given context. We present a tool \textsc{CitePrompt} which uses the hitherto unexplored approach of prompt learning for citation intent classification. We argue that with the proper choice of the pretrained language model, the prompt template, and the prompt verbalizer, we can not only get results that are better than or comparable to those obtained with the state-of-the-art methods but also do it with much less exterior information about the scientific document. We report state-of-the-art results on the ACL-ARC dataset, and also show significant improvement on the SciCite dataset over all baseline models except one. As suitably large labelled datasets for citation intent classification can be quite hard to find, in a first, we propose the conversion of this task to the few-shot and zero-shot settings. For the ACL-ARC dataset, we report a 53.86\% F1 score for the zero-shot setting, which improves to 63.61\% and 66.99\% for the 5-shot and 10-shot settings respectively.
\end{abstract}




\maketitle

\section{Introduction}
\begin{figure}[h]
\includegraphics[width=\linewidth]{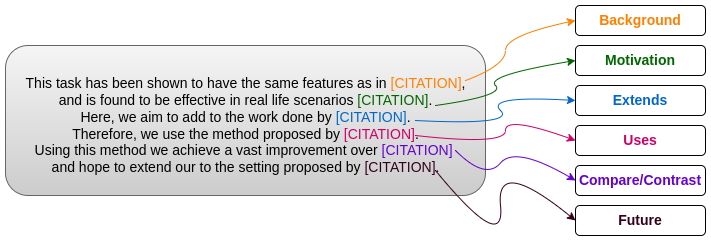}
\caption{Various citation intents labelled in a scientific text. The labels used here for illustration are from the annotation schema proposed for the ACL-ARC dataset \cite{jurgens-etal-2018-measuring}.} \label{fig1:citationIntent}
\end{figure}

Scientific works generally follow a policy of referring to prior work for a variety of purposes. The primary purpose is to situate the current work in the context of existing research. For example, citations may be used to trace or emphasize the motivation behind the problem, or to refer to a scientific resource that has been used in a research article, or to compare the results with other baselines in the given area. 
The kind of purpose a citation serves in a research work is referred to as its \textit{function} \cite{jurgens-etal-2018-measuring} or \textit{intent} \cite{cit_word_embedding}. Figure \ref{fig1:citationIntent} shows some of the intents associated with citations in a scientific article. 
\textit{Citation intent classification} helps in the large scale study of the behaviour of citations in scholarly digital libraries, such as the analytical study of citation frames on a temporal basis or exploring the effect of citation frames on future scientific uptake \cite{jurgens-etal-2018-measuring}.



Previous approaches to citation intent classification use rule-based or machine learning techniques, and in the latter category, the latest approaches use language models with `pre-train and fine-tune' paradigm \cite{kunnath2021meta}. In this paper, we use a different and novel technique based on prompt-based learning \cite{survey_prompt} to classify citation intents. Prompt-based methods use a fully natural language interface, allow quick prototyping, and are often useful in low-data regimes. 
Our work provides the following key contributions:
\begin{enumerate}
    \item We introduce a prompt-based learning tool, \textsc{CitePrompt}, for the task of automated citation intent classification.
    \item We achieve state-of-the-art results on this task for one dataset and comparable results for another.
    \item Our method runs only on the base task data and does not require additional training on scaffold tasks \cite{scicite}. Our method only uses an external scientific knowledge base for feeding external knowledge to the verbalizer.
    \item In a hitherto unexplored direction, we are the first to reformulate this task for the few-shot and zero-shot settings by leveraging the power of prompt engineering.
\end{enumerate}

\section{Related Work}
There has been a considerable amount of work in the field of citation intent classification although, to the best of our knowledge, neither the application of prompt engineering for citation classification nor few-shot citation intent classification has been attempted before. Prompt engineering is being increasingly applied in various text classification problems in zero-shot and few-shot settings \cite{lester-etal-2021-power, pet, hambardzumyan-etal-2021-warp}. Prompts are also used to interact with the conversational agent  {ChatGPT}\footnote{\url{https://openai.com/blog/chatgpt/}}.  
Study of the purpose of citations has been done since as early as the 1960s and 70s \cite{hirons, salton, Moravcsik_Murugesan}. Various automated approaches have been explored; examples include rule-based algorithm \cite{garzone}, nearest neighbour-based classification \cite{teufel-etal-2006-automatic}, support vector machine \cite{Agarwal2010AutomaticallyCT},  multinomial naïve Bayes \cite{Agarwal2010AutomaticallyCT}, and ensemble techniques \cite{dong-schafer-2011-ensemble}. Jurgens et. al. \cite{jurgens-etal-2018-measuring} not only create a new classifier and annotation schema that they apply on the NLP field, but also show how to use citation intents to gauge the general direction of research in a scientific field. Cohan et. al. \cite{scicite} are the first to introduce a multitask learning framework for citation intent classification along with a more coarse-grained annotation schema. Beltagy et. al. \cite{beltagy-etal-2019-scibert} consider the task of citation intent classification as a downstream task and so they finetune it on their SciBERT model. Mercier et. al. \cite{icaart21} propose an XLNet-based architecture for use in both citation intent classification and citation sentiment classification. For more details on the topic, the reader may consult the excellent contemporary surveys \cite{yousif2019survey,kunnath2021meta}.

\section{Proposed Method}\label{sec3}
Let the input be the citation sentence $x$ that is to be assigned a class label $y \in Y$, where $Y$ is the fixed set of labels for the citation intent classification task. An example of $Y$ is the set \{``Background", ``Motivation", ``Extends", ``Uses", ``Compare/Contrast", ``Future"\}.
Suppose $x = $ ``This task has been shown to have the same features as in [10]", then the desired label is $y=$ ``{Background}". 

We propose a framework called \textsc{CitePrompt} that uses prompt-based learning to solve the given task. The input citation sentence $x$ is first mapped by a \textit{template} $T$ to a prompt $x_T$: $T$ wraps $x$ and a \texttt{[MASK]} token into $x_T$. 
A pretrained language model $M$ is then fed with the string $x_T$ that outputs a distribution over its vocabulary for the \texttt{[MASK]} that acts as the answer slot. 
We construct a \textit{verbalizer} that chooses a set of label words $L$ from the vocabulary of the language model and maps them to the label space $Y$; 
more precisely, it maps a subset $L_y$ of $L$ to the label $y \in Y$. Given the output distribution from $M$, the distribution over every subset $L_y$ can be calculated, and hence, the label $y$ can be inferred to select the most probable label for the input $x$. 

In our model, we update the parameters of the language model $M$ as well as the verbalizer parameters to specify the model behavior. Our choice of fixed or manual prompts is motivated by the observation that as our task is precisely defined, we are able to create the prompt based on intuition rather than adopting an automated template strategy. Manual prompts have been used in GPT-2 \cite{NEURIPS2020_1457c0d6} and other zero-shot text classification problems \cite{0shot_tc}. The template used by us for citation intent classification is as follows,
\begin{center}
    \texttt{[X]} It has a citation of type \texttt{[MASK]}.
\end{center}
where \texttt{[X]} is the slot for an input citation sentence $x$. 
It is also called a \textit{prefix} prompt, distinct from a \textit{cloze} prompt where the slot to be filled is present somewhere in the middle of the given text. We found that our prefix prompt performs better than cloze prompts.

It is also quite necessary that we choose an appropriate answer space $L$ and create a mapping from this space to the original output space $Y$. 
Unlike many prompting methods, where the verbalizer makes use of only one or very few label words, we use a comparatively large set of such words such that these label words have an extensive coverage and little subjective bias. These two properties of label words are satisfied by selecting such words that belong to different facets and are intricately related to the label. We use an external knowledge base to choose the label words. In particular, for selecting the label words close to citation intent labels, we suggest choosing them from suitable text snippets where the label is most prevalent. Researchers \cite{teufel-etal-2006-automatic,dong-schafer-2011-ensemble,jurgens-etal-2018-measuring} have shown the strong dependence of the citation intent tag on the section the citation is present in. For example, the ``Background" citation label is present most frequently within the ``Introduction", ``Related Work", and ``Motivation" sections.  
Therefore, we propose the parsing of textual data in each of the major sections of each paper in a given collection of scientific papers to create a one-to-many mapping between the citation labels and the paper sections. The choice of sections for each label is mentioned in Table \ref{tab2}. Therefore, using this method for verbalizer construction, we try to capture the structure of the scientific document to better identify the citation purpose.

\begin{table}[h]
\centering
\caption{Table showing the sections where the corresponding label is most used. Please note that this table contains labels from both the ACL-ARC and the SciCite dataset, with ``Background" being the common label between the two.}\label{tab2}
\begin{tabular}{cc}
\toprule
\textbf{Label}     & \textbf{Corresponding paper sections}                        \\ 
\midrule
Background         & Introduction, Related Work, Motiv.                       \\ 
Method             & Methodology                                                  \\ 
Result             & Results                                                      \\ 
Motivation         & Introduction                                                 \\ 
Uses               & Motiv., Eval., Methodology, Results                 \\ 
Compares/Contrasts & Related Work, Results, Discussion                            \\ 
Extends            & Motiv., Methodology                                      \\ 
Future             & Conclusion, Discussion                                       \\ 
\bottomrule
\end{tabular}
\end{table}

More precisely, to construct the verbalizer, we first choose a few anchor words per label. Then we find the closest words to the anchor words. For example, the anchor words chosen by us for the label ``Method" are ``technique", ``procedure", and ``method". We find the closest words of an anchor word by computing the cosine similarity between the anchor word and the words parsed for the related sections (as given in Table \ref{tab2}) of papers in the given corpus of scientific documents. The union of all the resultant words for all the anchor tags serve as our label words for the corresponding label. This process helps us considerably broaden the label word space.

\section{Few-shot and Zero-shot Learning}\label{sec4}

A large volume of labeled citation data is not always available. We have incorporated some changes in our framework, \textsc{CitePrompt}, to adapt the citation intent classification task to low-resource settings.

\begin{figure}[h]
\includegraphics[width=\linewidth]{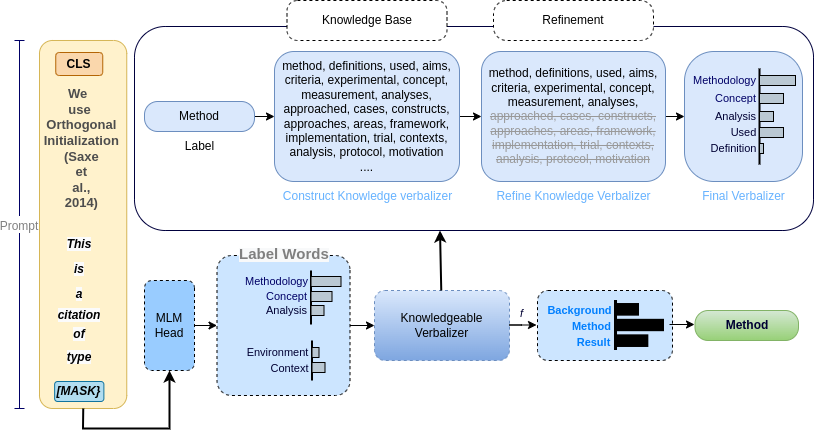}
\caption{Knowledgeable prompt-tuning (KPT) setup used for the few-shot and zero-shot settings in \textsc{CitePrompt}.} \label{fig2}
\end{figure}

We use the same verbalizer construction method as described above, except that we add the verbalizer refinement techniques from the Knowledgeable Prompt-tuning (KPT) framework \cite{hu-etal-2022-knowledgeable}. It comprises four optimization techniques that prove to be quite helpful in the few-shot and zero-shot settings as they help in reducing the noise present in the selected label word set. Originally, this noise gets introduced as the label word set was expanded in an unsupervised fashion during verbalizer construction. The noise reduction step becomes important because in the few-shot and zero-shot settings we do not have the luxury of training on many instances, and the label word noise affects the model performance. Very briefly, the refinement steps remove label words that are hard to find in the language model, remove irrelevant label words, calibrate the distribution predicted by the model, and assign a learnable weight to every label word \cite{hu-etal-2022-knowledgeable}.


Figure \ref{fig2} illustrates the KPT-augmented \textsc{CitePrompt} framework used by us. In the figure, the prompt containing the original input citation sentence and the \texttt{[MASK]} is passed through a masked language model (MLM), which outputs a distribution over the vocabulary, and then the \textit{knowledgeable verbalizer} converts it to a distribution over the citation intent labels, from which one of the labels is predicted as the output. Figure \ref{fig2} zooms into the \textit{knowledgeable  verbalizer} depicting that a label (here, ``Method") is first mapped to a set of label words drawn from an external knowledge base, and then the label word set is refined using the KPT techniques.

\section{Experimental Setting}\label{sec6}
\subsection{Dataset}
\label{sec5}
We use two standard citation intent classification datasets, \textbf{ACL-ARC} and \textbf{SciCite}.  
\textbf{ACL-ARC} \cite{jurgens-etal-2018-measuring} is  built from 186 papers from the ACL Anthology Reference Corpus \cite{bird-etal-2008-acl} and consists of 1,941 instances labeled with 6 citation intent labels. \textbf{SciCite} \cite{scicite} is a much larger dataset built from 6,627 papers and has 11,020 instances tagged with 3 categories of coarse-grained citation intents.

\subsection{Pretrained Language model}
We choose SciBERT \cite{beltagy-etal-2019-scibert} as our pretrained language model, which is a PLM trained on a corpus of 1.14M papers in  computer science and the biomedical domain. 

\subsection{Verbalizer Construction}
The S2ORC dataset \cite{lo-etal-2020-s2orc} gives us access to a huge corpus of parsed text data from scientific research papers, but not all the papers contain the sections that we listed below. Therefore, we select equal number of words for each section so that an equal distribution can be maintained. We select 100K words per section for each of the following sections from different papers present in the S2ORC dataset: Introduction, Related Work, Motivation, Methodology, Evaluation, Results, Discussion, and Conclusion. We use only a selection of Computer Science papers from S2ORC for this task.
Now that we have a large textual corpus for each section, we find the top words for each one based on the anchor words that have been selected for each section. We find 100 such keywords for each (anchor word, section) pair, which makes the total number of label words per label considerable for the verbalizer to work upon.

\subsection{Implementation Details}

We conduct our experiments using the OpenPrompt toolkit \cite{ding2021openprompt}, which is an open source framework for prompt learning. For all our experiments, we set the maximum sequence length to $512$ and the batch size to $40$. We ran every model for $5$ epochs. Across different settings and different datasets, we always report our results as the average of the results of $5$ different runs of the same experiment for $5$ seeds respectively. We conduct the experiments for the few-shot setting for 1-shot, 2-shot, 5-shot and 10-shot settings respectively. 
We report the F1-Macro and accuracy scores for every scenario.

\subsection{Baselines}

We consider the following models which have given high performance (including state-of-the-art results) as baselines. 
\textbf{Jurgens et. al. \cite{jurgens-etal-2018-measuring}} use a classifier based on various manually-selected features -- structural, lexical and grammatical, field and usage features -- that signify different aspects of a scientific paper. 
\textbf{Cohan et. al. \cite{scicite}} use a multitask learning framework containing BiLSTM with attention and utilizes ELMo embeddings \cite{peters1802deep}. They use two separate structural scaffold tasks to capture the scientific document structure. On top of these, they use a Multi-Layer Perceptron (MLP) for each task and then a softmax layer for obtaining the prediction probabilities. These layers contain task-specific parameters and are not shared among the different tasks.
\textbf{Beltagy et al. \cite{beltagy-etal-2019-scibert}} use SciBERT, a BERT-type model \cite{devlin2018bert} trained on scientific data and fine-tuned for several downstream scientific tasks including citation intent classification.
\textbf{Mercier et. al. \cite{icaart21}} use an approach based on XLNet, which is an auto-regressive language model containing bi-directional attention and is pretrained on a large amount of data. 
The last two baselines above have reported results only for the SciCite dataset.

\section{Results}\label{sec7}

\subsection{Fully Supervised Setting}

The results on the ACL-ARC dataset \cite{jurgens-etal-2018-measuring} are shown in Table \ref{tab-perf-ACL-ARC}. We observe that our method achieves an accuracy of 78.42\% and a macro-F1 score of 68.39\%, which indicates a clear increase in performance from the previous state-of-the-art models. The closest in performance is the method proposed by Cohan et. al. \cite{scicite}, but unlike them we do not make use of any additional tasks.   
\begin{table}[h]
\centering
\caption{Results for the ACL-ARC dataset.}\label{tab-perf-ACL-ARC}
\begin{tabular}{cccc}
\toprule
\textbf{Method} & \textbf{Dataset} & \textbf{Acc.} & \textbf{F1(Macro)} \\ 
\midrule
\begin{tabular}[c]{@{}c@{}}Jurgens et. al. \cite{jurgens-etal-2018-measuring} \end{tabular} & ACL-ARC          & NA                & 54.6               \\ 
\begin{tabular}[c]{@{}c@{}}Cohan et. al. \cite{scicite}\end{tabular}     & ACL-ARC          & NA                & 67.9               \\ 
\textsc{CitePrompt} (Ours)                                                                             & ACL-ARC          & \textbf{78.42}    & \textbf{68.39}     \\ 
\bottomrule
\end{tabular}
\end{table}

 The results on the SciCite dataset \cite{scicite} are shown in Table \ref{tab-perf-SciCite}. We achieve an accuracy of 87.56\% and a macro-F1 score of 86.33\%, which outperforms all of the baselines for this dataset except the present state-of-the-art score for this dataset.

\begin{table}[h]
\centering
\caption{Results for the SciCite dataset.}\label{tab-perf-SciCite}
\begin{tabular}{cccc}
\toprule
\textbf{Method}                                                                                         & \textbf{Dataset} & \textbf{Acc.} & \textbf{F1(Macro)} \\ 
\midrule
\begin{tabular}[c]{@{}c@{}}Jurgens et. al. \cite{jurgens-etal-2018-measuring} \end{tabular} & SciCite          & NA                & 79.6               \\
\begin{tabular}[c]{@{}c@{}}Cohan et. al. \cite{scicite} \end{tabular}     & SciCite          & NA                & 84.0                 \\
Beltagy et. al. \cite{beltagy-etal-2019-scibert}                                                                               & SciCite          & NA                & {85.49}              \\
Mercier et. al. \cite{icaart21}                                                                            & SciCite          & NA       & \textbf{88.93}     \\
\textsc{CitePrompt} (Ours)                                                                             & SciCite          & \textit{87.56}    & \textit{86.33}     \\ 
\bottomrule
\end{tabular}
\end{table}

\begin{figure}[h]
     \centering
     \begin{subfigure}[b]{0.45\linewidth}
         \centering
         \includegraphics[width=\linewidth]{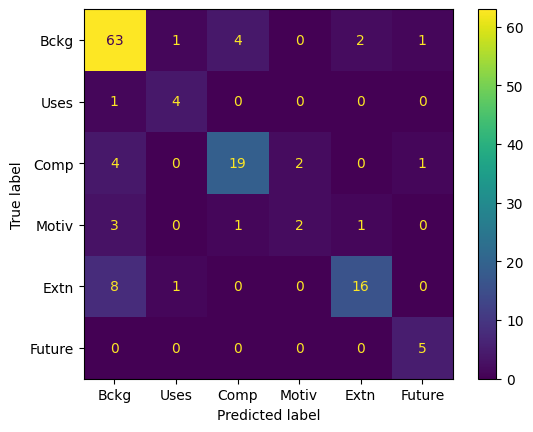}
         \caption{ACL-ARC}
         \label{fig3}
     \end{subfigure}
     \hfill
     \begin{subfigure}[b]{0.45\linewidth}
         \centering
         \includegraphics[width=\linewidth]{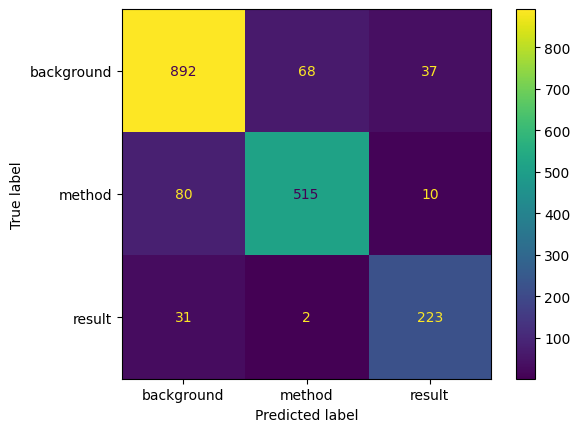}
         \caption{SciCite}
         \label{fig4}
     \end{subfigure}
        \caption{Confusion matrix showing classification errors of \textsc{CitePrompt} on two citation intent classification datasets.}
        \label{fig-confusion-matrix}
\end{figure}

The confusion matrix showing the nature of errors committed by our model is shown in Fig. \ref{fig-confusion-matrix}. In ACL-ARC, the errors are relatively few, with the model mostly making errors while identifying the ``Extends" label which it mislabels as ``Background" possibly because most of the times they both refer to an existing technique. Errors are more common in the SciCite dataset where instances with the true label ``Background" are sometimes misclassified as ``Method" (6.8\% times) and less frequently as ``Result" (3.7\% times).


\subsection{Few-shot and Zero-shot Settings}

We report the results from the SciCite dataset in zero-shot and few-shot settings in Table \ref{tab-perf-fs-SciCite}. We also tested for the ACL-ARC dataset, but it gave us very low results, the most likely reason for which may be attributed to the class imbalance present in the dataset.  
On SciCite, we are able to achieve an F1-score of 53.86$\%$ even in the zero-shot scenario.
There is a drop in performance of the model for the 1-shot scenario, but from that low point we see a steady increase in the performance of the model with increase in the number of shots both in terms of the F1 score and accuracy. As expected, we see a significant jump in performance in the 5-shot and 10-shot settings from the 1-shot and 2-shot settings. This shows that our model can perform citation intent classification with moderately good performance even for few-shot and zero-shot settings.

\begin{table}[h]
\centering
\caption{Results of citation intent classification on the SciCite dataset in the few-shot and zero-shot settings.}\label{tab-perf-fs-SciCite}
\begin{tabular}{cccccc}
\toprule
\textbf{Metric} & \textbf{Zero-shot} & \textbf{1-shot} & \textbf{2-shot} & \textbf{5-shot} & \textbf{10-shot} \\ 
\midrule
Accuracy        & 60.52             & 56.17           & 58.93           & 67.37           & 69.12            \\ 
F1(Macro)       & 53.86             & 48.66           & 51.81           & 63.61           & 66.99            \\ 
\bottomrule
\end{tabular}
\end{table}

\section{Discussion}

In the fully supervised setting, on both datasets, we clearly notice that our prompt-based model gives better results than the method used by Cohan et. al. \cite{scicite} and the automated feature-based citation classifier by Jurgens et. al. \cite{jurgens-etal-2018-measuring}. To better understand the structure of a scientific article, Cohan et. al. \cite{scicite} include two auxiliary scaffold tasks, which are the prediction of the title of the section where the citation occurs and the prediction of whether a sentence needs a citation or not. The F1 scores reported by Cohan et. al.'s method \cite{scicite} without the two scaffold tasks are 54.3 and 82.6 on the ACL-ARC and SciCite datasets respectively, which are significantly worse than our results. Similarly, Jurgens et. al. \cite{jurgens-etal-2018-measuring} use  structural features describing citation location, lexical and grammatical features representing citation description, and usage features representing external information.  Therefore, we are able to demonstrate that not only our method does not require multi-task training and requires less features, but also results in better performance. In our method, explicit feature engineering is not needed; the token representations are automatically learned through the use of SciBERT, and the information about the document structure that is needed to produce the final labels are obtained using an external knowledge base fed into the verbalizer.
We have transformed the classification task to a conditional text generation problem using prompt-based learning. Thereby, we have reformulated the problem to bring it closer to how humans approach such a task, which helps in solving it with significantly less external information than other approaches.

We observe from Table \ref{tab-ablation} that replacing SciBERT with BERT leads to a significant degradation of the performance. In particular, for  the ACL-ARC dataset, the accuracy drops by $4.75\%$ and the F1-score by $12.02\%$. Similar trend is observed for the SciCite dataset where the accuracy drops by $1.8\%$ and F1-score by $2.11\%$.

\begin{table}[h]
\centering
\caption{Effect of using SciBERT in \textsc{CitePrompt}.}\label{tab-ablation}
\begin{tabular}{cccc}
\toprule
\textbf{Method}                                                                                              & \textbf{Dataset} & \textbf{Acc.} & \textbf{F1(Macro)} \\ 
\midrule
\textsc{CitePrompt}     & ACL-ARC          & \textbf{78.42}    & \textbf{68.39}     \\ 
\textsc{CitePrompt} w/o SciBERT     & ACL-ARC          & 73.67             & 56.37              \\ \hline
\textsc{CitePrompt}     & SciCite          & \textbf{87.56}    & \textbf{86.33}     \\
\textsc{CitePrompt} w/o SciBERT     & SciCite          & 85.76             & 84.22              \\ 
\bottomrule
\end{tabular}
\end{table}

\section{Conclusion}\label{sec9}
We presented a new approach -- a prompt-based learning approach -- for citation intent classification, that is found to be effective in terms of performance and efficient in terms of extra training tasks required. Also, in a first, we show the effectiveness of this task in the few-shot and zero-shot settings. In the future, we aim to improve the performance further and incorporate multi-task learning in  our models to see if other similar tasks can enhance the performance of citation intent classification while using prompt engineering.

\begin{acks}
This work is supported by the SERB-DST Project CRG/2021/000803 sponsored by the Department of Science and Technology, Government of India at Indian Association for the Cultivation of Science.
\end{acks}

\bibliographystyle{ACM-Reference-Format}
\bibliography{sample-base}

\end{document}